# Median and Mode Ellipse Parameterization for Robust Contour Fitting

## Michael A. Greminger

**Abstract**— Problems that require the parameterization of closed contours arise frequently in computer vision applications. This article introduces a new curve parameterization algorithm that is able to fit a closed curve to a set of points while being robust to the presence of outliers and occlusions in the data. This robustness property makes this algorithm applicable to computer vision applications where misclassification of features may lead to outliers. The algorithm starts by fitting ellipses to numerous five point subsets from the source data. The closed curve is parameterized by determining the median perimeter of the set of ellipses. The resulting curve is not an ellipse, allowing arbitrary closed contours to be parameterized. The use of the modal perimeter rather than the median perimeter is also explored. A detailed comparison is made between the proposed curve fitting algorithm and existing robust ellipse fitting algorithms. Finally, the utility of the algorithm for computer vision applications is demonstrated through the parameterization of the boundary of fuel droplets during combustion. The performance of the proposed algorithm and the performance of existing algorithms are compared to a ground truth segmentation of the fuel droplet images, which demonstrates improved performance for both area quantification and edge deviation.

## 1 INTRODUCTION

Ellipse fitting to data in the presence of noise and outliers continues to be an area of active research. Ahn et al. proposed the least-squares orthogonal distance fitting method for fitting conics to data [1]. Ahn et al.'s method is robust to normally distributed noise but outliers impact the quality of fit. Fitzgibbon et al. developed an ellipse fitting technique that is robust to occlusion by restricting the fit to ellipses rather than all possible conics [2]. To increase the robustness of ellipse fits to outliers, methods have been developed to eliminate the outliers from the data. The most widely used method for removing outliers is the random sample consensus (RANSAC) approach to outlier rejection [3], which has also been applied to ellipse fitting by Duan et al. [4]. Other outlier rejection schemes have also been proposed such as the two-stage outlier rejection algorithm proposed by Yu et al. [5]. The Hough transform [6], which uses a voting approach to reject outliers, can be applied to ellipses as well [7].

An exhaustive five point ellipse fitting approach has been proposed by Rosin [8]. In Rosin's proposed approach, all combinations of five points from the data set are selected and a conic section is fit to each set of the five points. Conic sections that are not ellipses or circles are thrown out. The ellipse fit is then taken as the median parameter set (center, long axis length, short axis length, and angle) of all of the ellipses fit to the dataset. This approach performs well at rejecting outliers but has the limitation that an exhaustive ellipse fitting to all possible sets of five points in a source dataset requires $O(N^5)$ ellipse fits, where there are N points to be fit. Rosin addressed this problem by only using a uniformly spaced subset of all of the five point sets.

The curve fitting algorithm proposed here uses the multiple ellipse fitting approach of Rosin as the first stage of the algorithm. A subsampling strategy is used to limit the total number of ellipses that are fit to the data. The impact of this subsampling strategy will be characterized. This random sampling of ellipse fits is then used to parameterize the curve fit. The resulting fit is not actually an ellipse but is a median perimeter of all of the ellipses fit to the data. The unique property of the proposed algorithm is that it can fit closed contours that are more general than an ellipse while still robustly rejecting outlier points.

Finally, as a practical application of the proposed algorithm, the median ellipse parameterization algorithm will be applied to the problem of analyzing the combustion of fuel droplets.

## 2 MEDIAN ELLIPSE PARAMETERIZATION ALGORITHM

The first step in the median ellipse parameterization algorithm is to fit multiple ellipses to subsamples of five points from the source data set. One strategy would be to exhaustively choose all possible combinations of five points in the dataset. However, this would result in excessive computational cost for even a moderate number of data points so a random subsampling strategy will be used to fit only a fraction of the total number of five point combinations with ellipses. The impact of this ellipse subsampling will be quantified.

---

- M.A. Greminger is with the Department of Mechanical and Industrial Engineering, University of Minnesota Duluth, Duluth, MN 55812. E-mail: mgreming@d.umn.edu.

## 2.1 Ellipse Fitting

Five points is the minimum number of points required to uniquely fit an ellipse. Each ellipse can be expressed in the general conic form as

$$Ax^2 + Bxy + Cy^2 + Dx + Ey + F = 0 \qquad (1)$$

where the coefficients are determined by the following equation

$$\begin{vmatrix} x^2 & xy & y^2 & x & y & 1 \\ x_1^2 & x_1y_1 & y_1^2 & x_1 & y_1 & 1 \\ x_2^2 & x_2y_2 & y_2^2 & x_2 & y_2 & 1 \\ x_3^2 & x_3y_3 & y_3^2 & x_3 & y_3 & 1 \\ x_4^2 & x_4y_4 & y_4^2 & x_4 & y_4 & 1 \\ x_5^2 & x_5y_5 & y_5^2 & x_5 & y_5 & 1 \end{vmatrix} = 0 \qquad (2)$$

and $(x_i, y_i)$ are the five points that are used to define the ellipse. The conic section defined by (2) is not necessarily an ellipse unless it satisfies the following conditions [9]

$$B^2 2 - 4AC < 0 \qquad (3)$$

$$C\left(\left(AC - \frac{B^2}{4}\right)F + \frac{BED}{4} - \frac{CD^2}{4} - \frac{AE^2}{4}\right) < 0 \qquad (4)$$

A predetermined number of five point sets are randomly sampled from the list of points and only the five point sets that result in an ellipse are retained. Fig. 1 shows a set of ellipses fit to a set of 500 points. 100 conic sections were fit to 100 random samples of 5 points taken from the 500 original points. Fig. 1 shows 61 ellipses where the other 39 conics were not ellipses. In practice, 61 ellipses would not be enough to generate a good curve fit as will be seen below.

## 2.2 Median Curve Parameterization

Once the desired number of ellipses are fit to the data, the curve can now be parameterized. A polar parameterization is used for the curve. The first step in the parameterization is to locate the center of the curve fit. This is done in the manner suggested by Rosin [8] where the median center of the all of the ellipse fits is taken as the center for the polar parameterization of the curve fit. Using this center point, a ray is cast for a given angle $\theta$. The intersections between this ray and each of the ellipses that were fit to the data are computed. Only ellipses that enclose the center point are used, which eliminates cases of multiple or no intersections. The radius of each intersection is computed and the median radius of all of the intersections is taken to be $r$ for the current value of $\theta$. Fig. 2 shows one example of the computation of the median ellipse intersection point. Only a small number of ellipses are shown in this figure for clarity. Using the median intersection, rather than the average intersection, enables this algorithm to be robust to outliers.

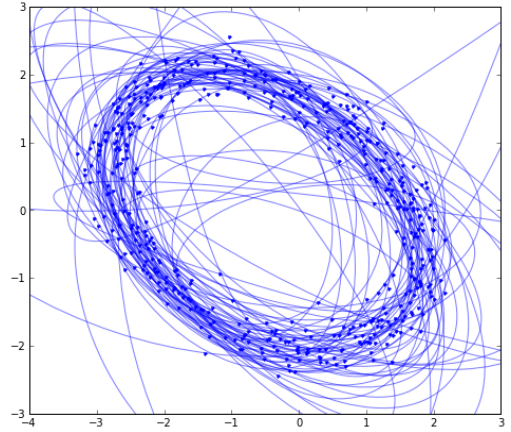

Fig. 1. Ellipses fitted to subsets of five points

Since the median point is chosen independently for every ray angle $\theta$, a different ellipse may be chosen as the median point at different locations along the curve fit. Because of this property of the median ellipse parameterization algorithm, the resulting curve fit is not an ellipse. However, it is desirable for the median ellipse parameterization to degenerate to an ellipse when the underlying data represents an ellipse as will be illustrated below.

The median ellipse parameterization is guaranteed to have $C^0$ continuity since the median point will remain on a single ellipse over a range of $\theta$ until another ellipse crosses that ellipse causing the median ellipse fit to move to the next ellipse at the crossing point. However, $C^1$ continuity is not guaranteed since the crossing ellipses will have different slopes. Even though $C^1$ continuity is not enforced, it will be seen from the results that the median ellipse parameterizations are smooth due to the large number of ellipses used to generate the median ellipse parameterization.

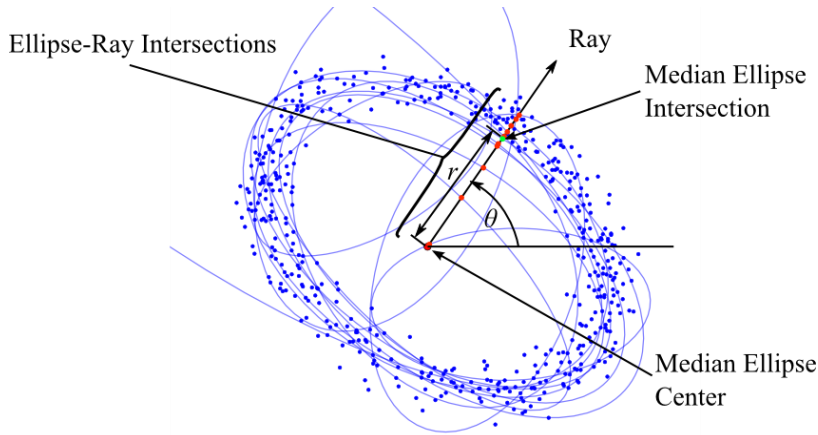

Fig. 2. Calculation of curve fit radius based on the median intersection of the ellipse fits.

## 3 MEDIAN ELLIPSE PARAMETERIZATION APPLIED TO NOISY DATA WITH AND WITHOUT OUTLIERS

The simplest case to test the proposed algorithm is with source data that represents an ellipse with normally distributed noise. The main purpose of this test case is to insure that the median ellipse parameterization algorithm will degenerate to an ellipse if the underlying curve is an ellipse. Fig. 3 shows the noisy ellipse dataset along with an orthogonal least squares fit to the dataset based on the algorithm proposed by Ahn et al. [1]. Fig. 3 also shows the median ellipse parameterization fit to the same noisy ellipse dataset. From Fig. 3, it can be seen that median ellipse parameterization algorithm performs similarly to the orthogonal least squares algorithm for the case of an ellipse with normally distributed noise. To generate the median ellipse parameterization, 10000 conics were fit to the data

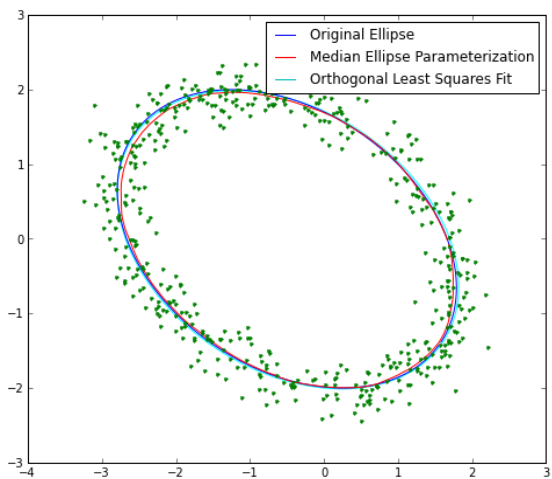

Fig. 3. Noisy ellipse shown with original ellipse, orthogonal least squares fit, and the median ellipse fit.

with 5747 of those conics being ellipses (Fig. 4 shows a cloud of these 5747 ellipses).

Since only a subsample of the total number of possible conic fits are being used to generate the median ellipse parameterization, it is important to characterize the impact that this subsampling has on the fit accuracy. Over $2 \times 10^{11}$ conic fits would be required to exhaustive fit all combinations of five points to a 500 point dataset. To perform this number of fits would be computationally prohibitive so only a small fraction of the total number of possible conic fits are used. Fig. 5 shows the influence of the number of ellipses on the quality of the fit. The error is quantified in Table 1 where the error to the true ellipse is averaged over

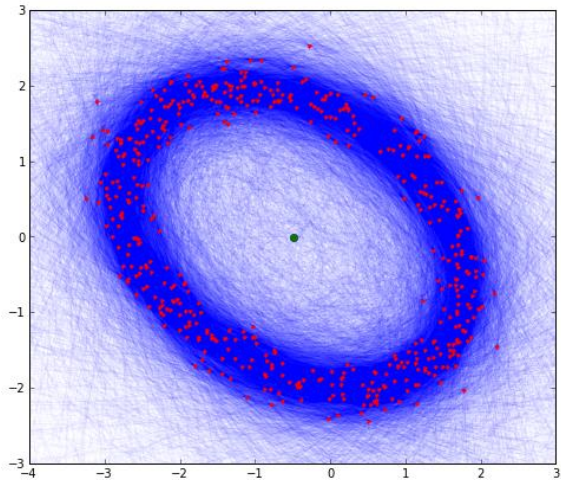

Fig. 4. Ellipse cloud used to generate the median ellipse parameterization curve fit shown in Fig. 3.

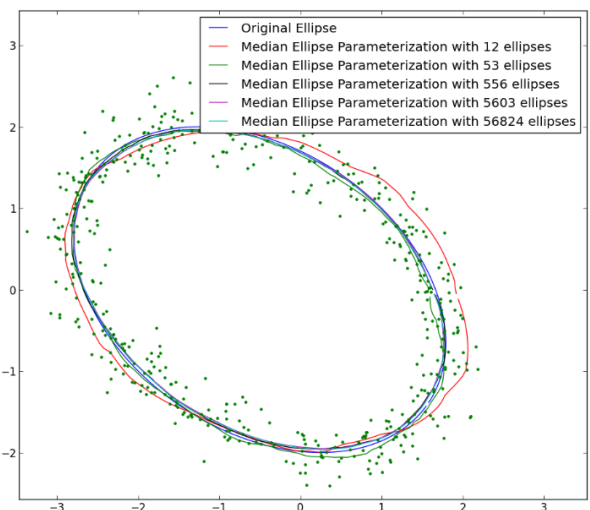

Fig. 5. Impact of the number of ellipses on the median curve fit.

200 points around the circumference of the median curve fit. It can be seen from the table that for 556 or more ellipses, the error levels off to a magnitude that is only slightly higher than was obtained using an orthogonal least squares fit. This verifies that a subsampling of the possible ellipse fits can be used to generate a curve parameterization with accuracy similar to a least squares fit for the case of a noisy ellipse. When median ellipse parameterization is applied to a new problem, a similar convergence study should be performed in order to insure that a sufficient number of conics are used to generate the median curve fit.

Table 1. Conic fit quantity impact on the mean error.

| Number of Conics Fit to Data | Number of Conics that are Ellipses | Mean Error (200 points) |
|---|---|---|
| 20 | 12 | 0.116 |
| 100 | 53 | 0.038 |
| 1000 | 556 | 0.027 |
| 10000 | 5603 | 0.031 |
| 100000 | 56824 | 0.030 |
| Mean Error for Orthogonal Least Squares Fit | | 0.027 |

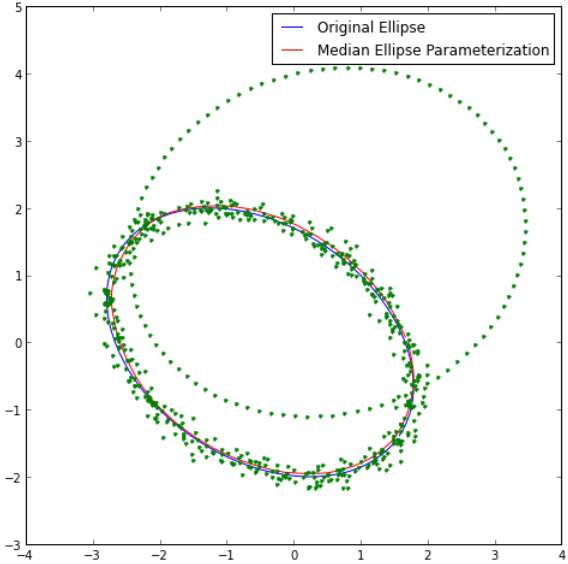

Fig. 6. Median ellipse parameterization applied to an example with outliers.

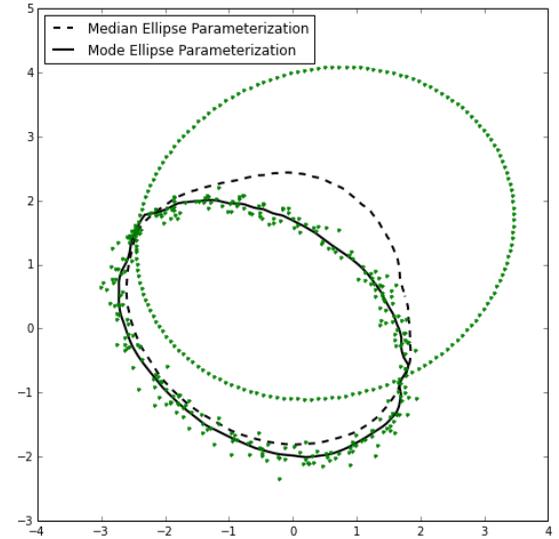

Fig. 7. Median ellipse parameterization applied to an example with an increased number of outliers.

Fig. 6 shows an example of a median ellipse parameterization curve fit to a noisy ellipse in the presence of outliers. The noisy ellipse consists of 500 points and there are 100 outliers from a different ellipse. Again, 10000 conics were fit to the data with 5209 being ellipses. It can be seen from Fig. 6 that the original ellipse is fit well even in the presence of a significant number of outliers which themselves are from a different ellipse. This robustness property of the median ellipse parameterization algorithm will be used below to analyze fuel droplet combustion images.

## 4 IMPACT OF MULTIMODALITY ON THE PERFORMANCE OF MEDIAN ELLIPSE PARAMETERIZATION

As has been shown, the median ellipse intersection for the parameterization of a curve fit can handle some level of outliers in the data. However, if the distribution of ellipse intersections for each ray becomes multimodal, the median statistic will begin to fail, especially if the mode peaks in the distribution start to become comparable in size. Fig. 7 shows the case where, if the number of outliers is increased to 200 points, the median ellipse parameterization fails. Using the mode rather than the median can be beneficial for cases where the distribution is multimodal. The mode for each ray is defined as the radius with the most ellipse intersections. This can be computed using a histogram of the ellipse intersection radii for each ray. However, the results using a histogram are highly dependent on the chosen bin locations and the resolution is limited by the bin width. Some of the limitations of using a histogram can be alleviated by using kernel density estimation (KDE) to estimate the probability density function of the ellipse intersection distribution [10]. A Gaussian kernel was used for the KDE results shown here. In addition, a bandwidth needs to be chosen to scale the width of the kernel. The bandwidth is analogous to the bin width for a histogram. Fig. 7 shows that the mode ellipse parameterization performs well for the larger number of outliers that causes the median ellipse parameterization to fail. Fig. 8 shows the KDE for one particular ray used to generate Fig. 7.

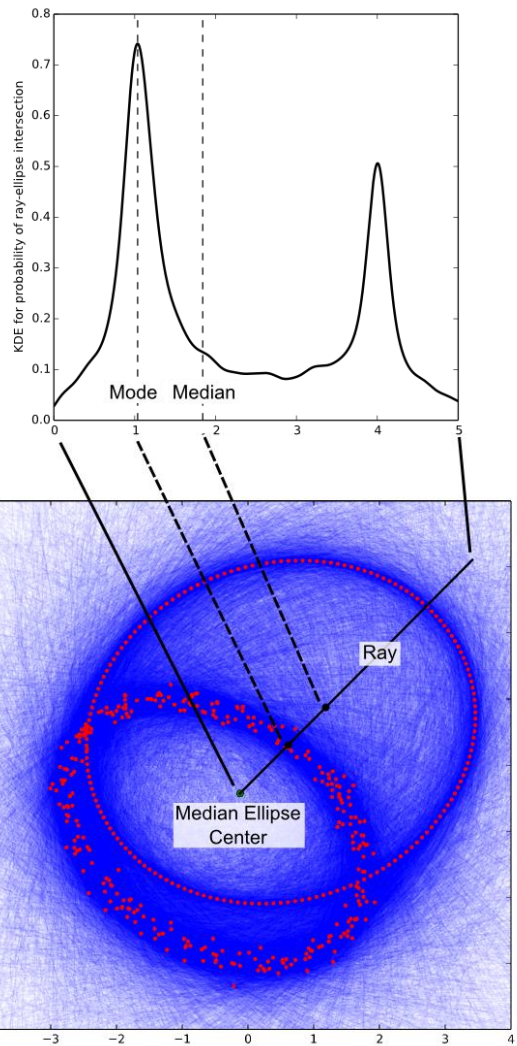

Fig. 8. The ellipse cloud used to generate the fit shown in Fig. 7 is shown at the bottom. The kernel density estimation along a single ray is shown at the top with the median and mode locations indicated by vertical lines.

As discussed above, $C^0$ can be guaranteed with median ellipse parameterization. However, no such guarantee can be made with mode ellipse parameterization since nothing prevents the peak of the KDE of adjacent rays to swap between peaks. As will be seen below, the insensitivity to multimodality of the mode ellipse parameterization algorithm can provide a performance benefit over median ellipse parameterization for some cases. However, the drawback of the lack of enforced continuity needs to be weighed against these performance benefits for each potential application.

## 5 QUANTITATIVE COMPARISON OF ROBUST ELLIPSE FITTING ALGORITHMS

### 5.1 Implementation of the Methods Compared

The dominant robust curve fitting algorithms can all be expressed in a framework, where, similar to the proposed algorithm and the Rosin median ellipse algorithm [8], the first step is to fit a fixed number of ellipses to randomly selected subsamples of the data points. The original RANSAC algorithm [3] used this approach to fit a model to data. In the RANSAC algorithm, the curve fit from the randomly sampled points that has the most inlier points is chosen as the best fit. An orthogonal least squares fit is then made to the inlier points from this best fit. A RANSAC implementation based on the original algorithm has been implemented for the comparisons that follows.

The Rosin median ellipse algorithm was implemented according to the original algorithm [8] with one change. In the Rosin algorithm, the median ellipse parameters were obtained by calculating the median of the natural ellipse parameterization (center x, center y, major axis length, minor axis length, and major axis angle) rather than using the general conic coefficients given in (1). Rosin indicated that using the natural parameters gives better results. However, the one issue that Rosin did not address is the special attention that has to be taken when performing a mean or median calculation on a directional quantity such as the major axis angle in the natural parameters for an ellipse. The issue is that for angles defined on the interval $[-\pi/2, \pi/2]$, a major axis angle of $\pi/2$ is identical to a major axis angle of $-\pi/2$ because of the periodic nature of the angle parameter and the symmetry of an ellipse. A standard median calculation would not consider the angles of $\pi/2$ and $-\pi/2$ to be near each other. For periodic distributions such as this, the median for a directional statistic is define as the angle the splits the main cluster of angles in two equal halves [13]. The median for a directional statistics is only well defined if there is a single main cluster of directions. This approach may break down if there are many ellipses in the image.

The original Hough transform algorithm [6][14] relied on fitting lines to all points in the image and then accumulating the two parameters that define each of the lines in a histogram. The randomized Hough transform (RHT) introduced by Xu et al. [11] changed the original Hough transform by fitting lines to random subsamples of point pairs in the original image. In that work, Xu et al. also devised an algorithm to implement a sparse accumulator space to alleviate the need to maintain a histogram in the parameter space for the curve to be fit. Because of the sparse accumulator, the RHT algorithm is more suited to fitting curves with more parameters then a line, such as is the case for circles and ellipses. Xu et al. applied their RHT algorithm to lines and circles and McLaughlin [12] expanded the algorithm to the fitting of ellipses.

The RHT ellipse fitting algorithm implemented for this comparison is based on that of McLaughlin but differs in two ways. McLaughlin fit ellipses to sets of three points and their slopes, where the algorithm implemented for the comparison uses the five-point ellipse sampling approach discussed above. Second, a kernel density estimation approach [10] with a Gaussian kernel was used to find the most popular ellipse in the parameter space rather than the sparse accumulator used in the RHT algorithm. The change to a kernel density estimation voting procedure was done merely to simplify the implementation since the sparse accumulator proposed in the original RHT algorithm effectively implements a kernel density estimator with a uniform kernel. Like the Rosin implementation, McLaughlin recommends using the natural ellipse parameters in the accumulator space. This raises the same issue with the major axis angle parameter as with the Rosin median implementation discussed above. To alleviate this issue, the directional median of the angles as taken and subtracted from all of the source ellipses before performing the RHT procedure so that the ellipse angles will be centered about an angle of zero which is in the center of the interval used for the major axis angle: $[-\pi/2, \pi/2]$. The median angle is than added back to the result. This approach works well when there is a single ellipse that is being fit. If multiple ellipses need to be fit, which is one of the benefits of the Hough transformation, a kernel density estimator that accounts for the periodicity of the major axis angle would have to be implemented. Multiple ellipse fitting for the RHT was not implemented for the comparison presented here but it is important to point out that the ability to fit more than one ellipse is an advantage of the RHT ellipse fitting algorithm.

Table 2: Summary of algorithms compared.

| Ellipse Fitting Algorithm | Required Input Parameters | Output | $C^0$ Continuity? | Fit Multiple Ellipses? |
|---|---|---|---|---|
| **Orthogonal Least Squares** [1] | 1. Initial Guess | Ellipse | Yes | No |
| **RANSAC** [3], [4] | 1. Random Ellipse Count<br>2. Inlier Threshold | Ellipse | Yes | No |
| **Randomized Hough Transform** [11], [12] | 1. Random Ellipse Count<br>2. KDE Bandwidth | Ellipse | Yes | Yes |
| **Rosin Median Ellipse** [8] | 1. Random Ellipse Count<br>2. KDE Bandwidth | Ellipse | Yes | No |
| **Median Ellipse Parameterization** | 1. Random Ellipse Count | Polar Paramaterization | Yes | No |
| **Mode Ellipse Parameterization** | 1. Random Ellipse Count<br>2. KDE Bandwidth | Polar Paramaterization | No | No |

The algorithms that were compared are the presently proposed algorithms of median and mode ellipse parameterization, the median ellipse parameterization algorithm proposed by Rosin, the randomized Hough transform implemented for ellipses, the RANSAC algorithm implemented for ellipses, and orthogonal least squares ellipse fitting.

These algorithms are summarized in Table 2. All of the algorithms, with the exception of orthogonal least squares algorithm, require a list of ellipses fit to random samples of the data as a starting point. The same ellipse list was provided for all of the algorithms in order to eliminate any variability due solely to the ellipse sampling. All of the algorithms were implemented in the Python programming language. The Scikit-Learn library [15] was used for kernel density estimation and the SciPy [16] and NumPy [17] Python libraries were used for low level image processing and numerical computations.

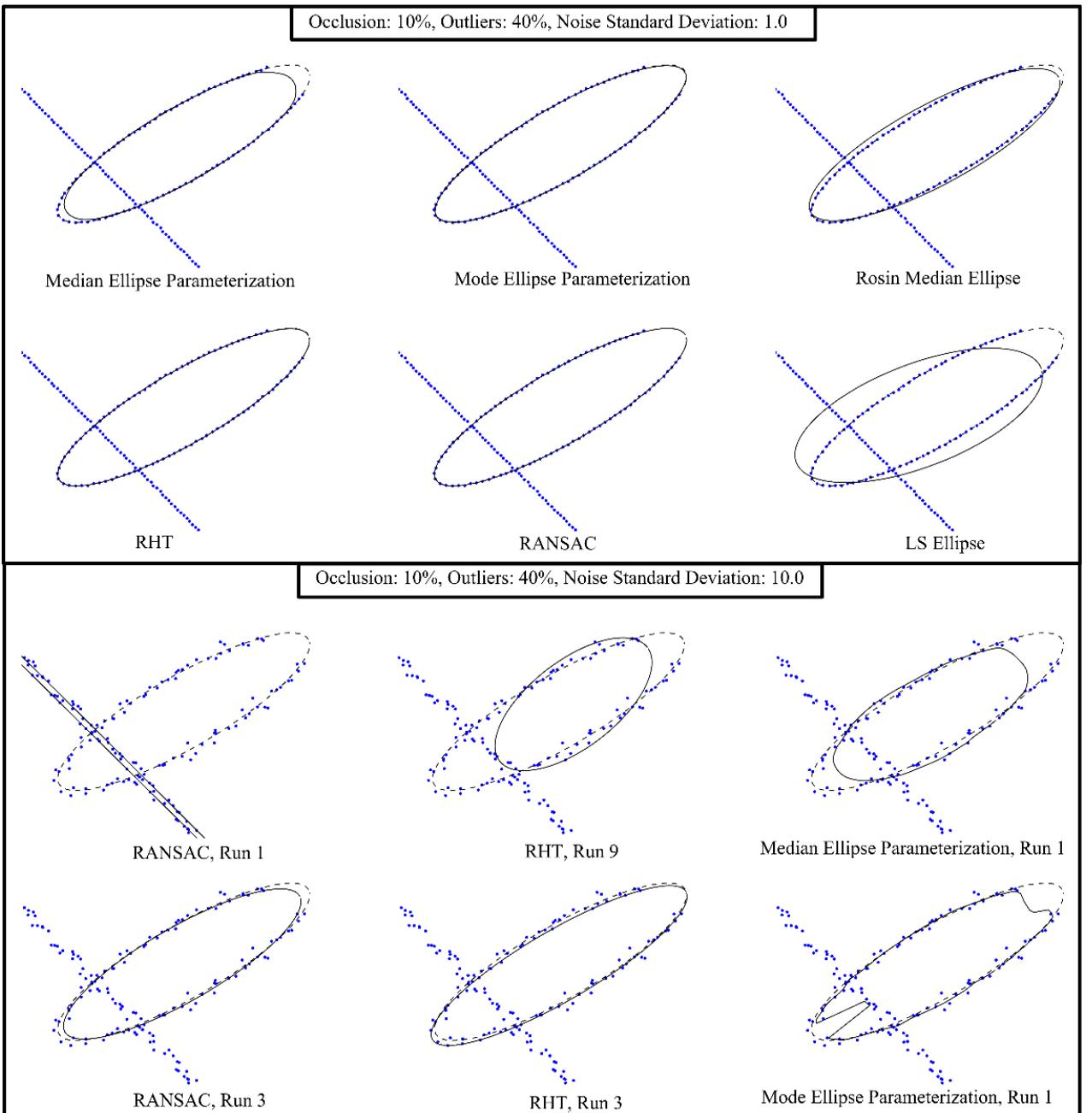

Fig. 9: Selected results from ellipse fitting comparison study.

Table 3: Average error for each of the algorithms compared.

| Number Ellipses | Outlier Ratio | Occlusion Ratio | Noise Sigma | LS Ellipse Error | Median Ellipse Parameterization Error | Mode Ellipse Parameterization Error | RANSAC Ellipse Error | RHT Ellipse Error | Rosin Median Ellipse Error |
|---|---|---|---|---|---|---|---|---|---|
| 10000 | 0.1 | 0.1 | 1 | 11.09 | 0.38 | 0.34 | 0.25 | 0.25 | 0.45 |
| | | | 10 | 11.81 | 4.76 | 4.28 | 5.30 | 10.96 | 7.14 |
| | | 0.4 | 1 | 18.33 | 1.36 | 1.01 | 0.63 | 1.31 | 3.16 |
| | | | 10 | 19.80 | 10.58 | 10.81 | 7.16 | 15.03 | 18.59 |
| | 0.4 | 0.1 | 1 | 39.56 | 2.74 | 0.43 | 0.16 | 0.22 | 9.29 |
| | | | 10 | 37.54 | 13.62 | 6.74 | 5.77 | 20.28 | 30.88 |
| | | 0.4 | 1 | 10328.24 | 5.85 | 3.75 | 0.53 | 1.00 | 15.69 |
| | | | 10 | 16331.34 | 16.67 | 12.19 | 9.89 | 25.14 | 35.22 |

## 5.2 Comparison of Results

The factors that were varied when comparing the algorithms were the number of number of random ellipse samples, percentage of outlier points, percentage of occluded points, and the standard deviation of the Gaussian noise applied to the test points. Fig. 9 shows a small number of the runs. The tested algorithms (except for the least squares algorithm) are inherently non-deterministic since a random subsample of ellipses are used for each algorithm. To account for the variability due solely to random ellipse sampling, 10 runs were made for each algorithm per set of input points. The error for each algorithm was quantified by computing the average distance from 200 points on the ellipse fit to the actual ellipse (the actual ellipse is shown as a dashed line in Fig. 9). 24 total input cases were run 10 times each resulting in 240 total runs. The median of the average errors for each of the 10 runs for each configuration are summarized in Table 3. The four algorithms with the lowest error for each case are highlighted in the table. The method with the least error is given the darkest highlight.

It can be seen from Table 3 that the RANSAC algorithm has the lowest error for almost all cases. The second best algorithm is either the RHT algorithm or the mode ellipse parameterization algorithm (in one case, the median ellipse parameterization algorithm had the second best performance). In all cases with the higher noise standard deviation, the median or mode parameterization algorithms perform better than the RHT algorithm. The median and mode ellipse parameterization algorithms have nearly identical performance for all of the cases with 10% outliers. However, for the cases with 40% outlier points, the mode ellipse parameterization algorithm performs significantly better than the median ellipse parameterization, which follows directly from the discussion above on the motivation for the mode ellipse parameterization for multimodal data sets. A case where the mode ellipse parameterization leads to a non-continuous parameterization can be seen in the lower right result shown in Fig. 9. Fig. 9 shows some results for the case where there are 40% outlier points, 10% occluded points, and a noise standard deviation of 1 for the top half of the figure and a noise standard deviation of 10 for the bottom half of the figure. As was mentioned above, each algorithm was run 10 times for each case. Fig. 10 shows the histogram for the 10 runs for the input points shown in the bottom half of Fig. 9. The median parameterization and mode parameterization algorithms have errors that change very little from run to run. However, both the RHT and RANSAC algorithms have error that varies widely from run to run, including an occasional outlier with very large error (two of these outlier cases are shown in the bottom half of Fig. 9). This non-normal distribution of error is why the median of the ten runs is reported in the error summary given in Table 3.

The algorithms can also be differentiated by their run times. Table 4 summarizes the average run time versus number of ellipse for all of the cases that were run. The times in Table 4 include the time required to generate the ellipse samples. The RHT and Rosin median algorithms are the fastest algorithms. In general, the RANSAC algorithm is the slowest algorithm. However, for the 10000 ellipse case, the mode ellipse parameterization algorithm is the slowest algorithm. The mode ellipse parameterization algorithm scales poorly with large numbers of ellipses. The KDE estimation is the bottleneck for the mode ellipse parameterization algorithm when there are a large number of ellipse intersections for each ray. One caveat on the time results is that the algorithms were implemented in Python without extensive consideration of the efficiencies of the implementations. However, it is believed that the time results provide some insight into how each of the algorithms scales with an increasing number of ellipses.

Table 4: Run time for the robust ellipse fitting routines evaluated.

| | Number of Ellipses | | |
|---|---|---|---|
| | 100 | 1000 | 10000 |
| Median Ellipse Parameterization time (sec) | 2.5 | 23.3 | 230.6 |
| Mode Ellipse Parameterization time (sec) | 4.1 | 36.5 | 1400.0 |
| RANSAC Ellipse time (sec) | 12.0 | 45.1 | 380.2 |
| RHT Ellipse time (sec) | 1.4 | 2.9 | 52.6 |
| Rosin Median Ellipse time (sec) | 1.4 | 2.9 | 49.1 |

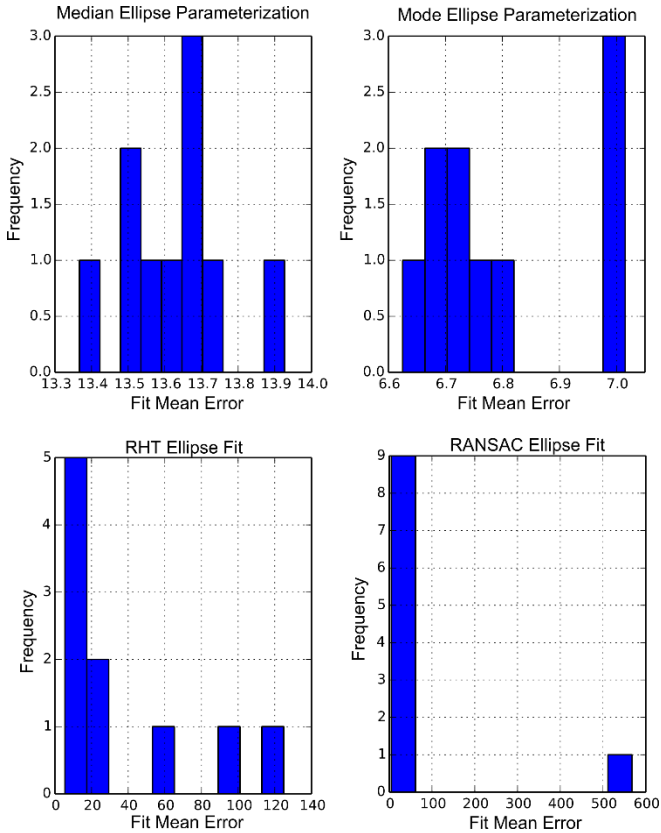

Fig. 10. Histogram of errors for 10 runs of the robust ellipse fitting algorithms. Plotted for the case with 40% outliers, 10% occlusions, and high noise.

The final information that will be summarized from this comparison is the influence of the number of randomly sampled ellipses used on the error of the fits. Fig. 11 shows the fit error versus number of randomly sampled ellipses. The errors are normalized to the error using 100 ellipses. The general trend for all of the algorithms is for the error to decrease as the number of ellipses is increased. However, some algorithms are more sensitive to this trend. The algorithms that are based on the median statistic, median ellipse parameterization and Rosin median ellipse, are less sensitive to the number of ellipses then the algorithms that depend on estimating the probability density function using the KDE approach, specifically, mode ellipse parameterization and RHT. The median statistic seems to be less sensitive to the number of samples as compared to the mode statistic. This is likely due to that the mode being obtained from the KDE estimate of the probability density function. The KDE converges to the true probability density function as the number of samples is increased. In addition, the RANSAC algorithm falls into the category of algorithms that are more sensitive to the total number of ellipses. This is likely because the presence of more ellipses increases the probability of finding an ellipse with an improved fit.

## 6 MEDIAN ELLIPSE PARAMETERIZATION APPLIED TO THE ANALYSIS OF FUEL DROPLETS

One property of the median ellipse parameterization algorithm is that it can represent shapes that are more general than an ellipse. There are many examples in computer vision where ellipse-like shapes need to be extracted from an image. The example that will be used here to quantify the performance of the proposed algorithm is the analysis of the combustion of fuel droplets. Hoxie et al. [18] has used an optical analysis of droplet size over time to characterize the combustion of butanol-soybean oil fuel blends. During some portions of the combustion, the droplet is stable and an ellipse fit to the droplet is an accurate predictor of droplet area. However, at various points during the combustion, microexplosions occur that cause erratic changes in the droplet shape. During these periods of microexplosions, an elliptical fit is not adequate to quantify the shape and size of the droplet.

A set of these droplet combustion images will be analyzed to quantify the performance of the proposed median and mode ellipse parameterization algorithms. The algorithms will be compared to the RHT and RANSAC algorithms.

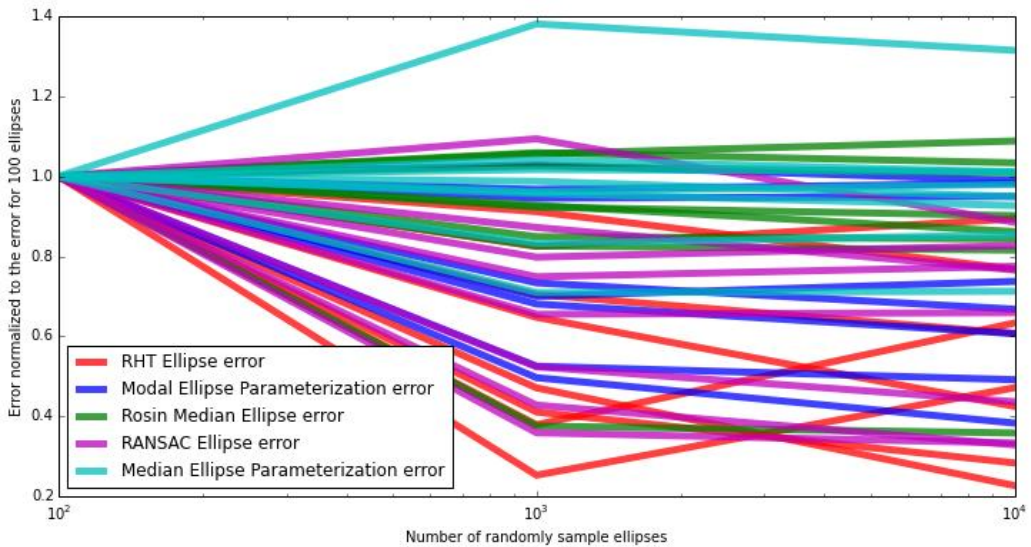

Fig. 11: Error versus number of randomly samples ellipses, normalized to the error for 100 ellipses.

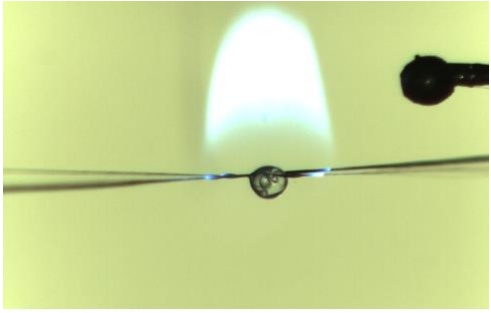

Fig. 12: Burning droplet of fuel.

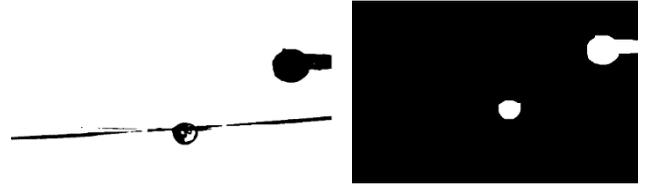

Fig. 13: Otsu threshold image on left. Image after morphological operations on right.

## 6.1 Edge Point Extraction Algorithm

Fig. 12 shows one frame from the droplet combustion image sequence. The first step in analyzing the images is to identify the location of the droplet and to take a region of interest centered around the droplet. The image is first converted to a binary image using Otsu's method [19], which is shown on the left half of Fig. 13. Next, the wires from which the droplet are suspended are removed from the threshold image by a specific sequence of morphological dilation and erosion operations. The resulting binary image is shown on the right half of Fig. 13.

The only objects that remain in the binary image are the droplet to be analyzed and the thermocouple used to monitor combustion temperature in the upper right corner of the image. The region closest to the center of the image is taken to be the droplet region. Next, the Canny edge operator [20] was applied to the original image in the region of the droplet (see the left half of Fig. 14). Finally, the edge points furthest from the droplet center in the radial direction are taken as the points that will be fit by the various algorithms (see the right half of Fig. 14). It can be seen from the right half of Fig. 14 that most of the edges have been classified correctly as the boundary of the droplet. However, at locations where the wires intersect the droplet, there are several spurious edges along with some missing edges. This misclassification of some of the edges requires that a robust algorithm is used to find the contour of the droplet.

## 6.2 Curve Fitting Results

One quantity of interest when analyzing the droplet combustion is the area of the droplet. A manual segmentation was performed on each of the 24 test images to act as the ground truth for calculating errors. One method to automatically obtain the area is to simply calculate the area of the region found using the morphological operations described above. In addition to the morphological area, the median ellipse parameterization, mode ellipse parameterization, RHT, and RANSAC algorithms were applied to the droplet combustion images.

The above procedure for finding the edge points on the boundary of the droplet was repeated for a subset of 24 images from the droplet image sequence. The curve fits, along with the manual segmentation, are shown for three of the images in Fig. 15. During the initial phases of combustion, the droplet is approximately elliptical and all of the algorithms have similar performance and nearly match the manual segmentation (see the left image in Fig. 15). However, once the microexplosions begin, the droplet shape is no longer elliptical and the ellipse based algorithms perform poorly (see the center and right images in Fig. 15). For these microexplosion images, the mode ellipse parameterization performs better than the median ellipse parameterization algorithm. Table 5 summarizes the average error in the area calculation for the four curve fitting algorithms and the morphological approach to area calculation. The median and mode ellipse parameterization algorithms have the lowest error when averaged across all of the images. Table 5 also quantifies the error when calculated as a radial edge deviation from the curve fit to the manual segmentation. Again, the median and mode ellipse parameterization algorithms have the lowest average edge deviation when averaged across all 24 images. These algorithms have edge deviation errors that are about half of those obtained when using the RANSAC or RHT algorithms.

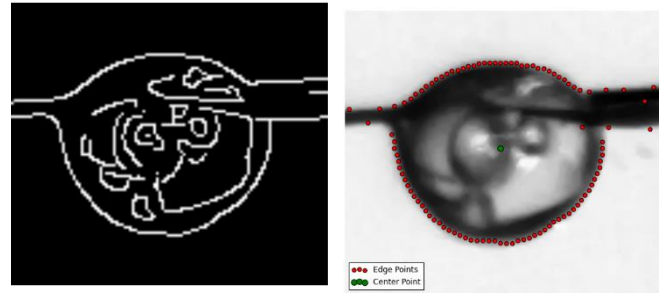

Fig. 14: Canny edge image on left and edge points used for curve fitting on right.

Fig. 16 plots both the area error and the average edge deviation versus image number for all of the images in the sequence. The images for which the RANSAC and RHT algorithms have large errors correspond to the images where microexplosions are occurring (for two examples, see the center and right hand images in Fig. 15).

If area is all that is needed from the analysis of the droplet combustion images, then the morphological approach has fairly good performance. However, it is also useful to quantify the duration and severity of the microexplosions, which can only be done when the shape of the droplet is known. It has been shown that the median or mode ellipse parameterization algorithms provide a significant improvement on the characterization of the contour of the droplets when compared to existing algorithms.

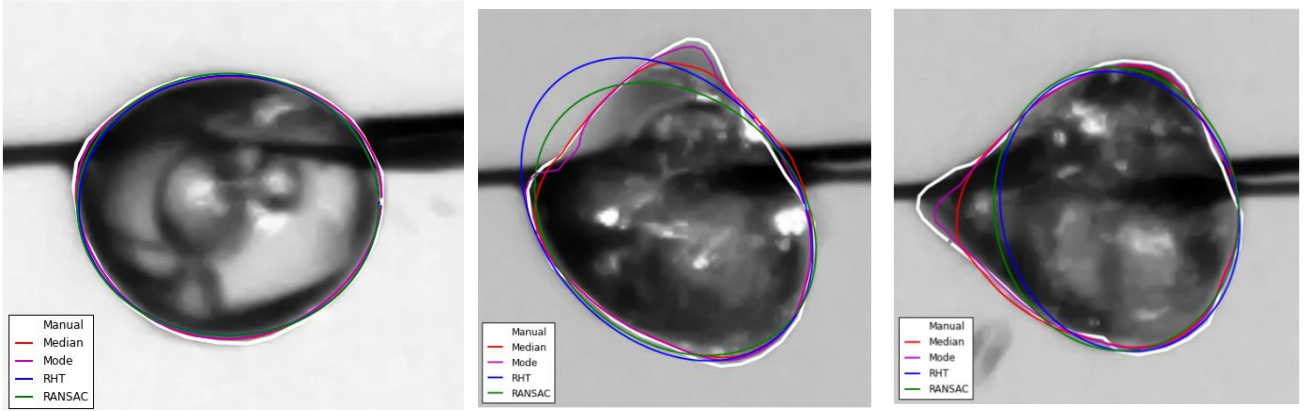

Fig. 15: Three frames from the droplet combustion image sequence with curve fits. Droplet before microexplosions on the left and droplet after the initiation of microexplosions at the center and right.

## 7 CONCLUSIONS

Both a median and mode ellipse parameterization algorithm were introduced that allow the robust parameterization of closed contours with both outliers and occlusions. It was shown that median ellipse parameterization performs well for cases where there are few outliers and the underlying data to be fit is relatively smooth. In addition, it was show that the median parameterization guarantees $C^0$ continuity of the resulting parameterization. Alternatively, the mode parameterization algorithm perfoms well in the cases with a significant number of outliers or when the underlying curve to be fit is less smooth. The median parameterization approach does have the drawback of not guaranteeing $C^0$ continuity of the parameterization.

It was shown that, if the contours to be fit represent an ellipse and when there is significant noise, the resulting, median and mode parameterizations perform better than the randomized Hough transform but worse than the RANSAC algorithm. In terms of computational efficiency, the median ellipse parameterization is more computationally efficient than the RANSAC algorithm but is more computationally expensive than the RHT algorithm. It was also shown that the median and mode ellipse parameterization algorithm can represent shapes that are more general than an ellipse. This was demonstrated by parameterizing the boundary of a fuel droplet during combustion. For the analysis of the fuel combustion droplet boundaries, it was shown that both the median and mode ellipse parameterization algorithms performed better than both the RANSAC and RHT algorithms for both droplet area and average edge deviation.

Median and mode ellipse parameterization provide a general set of algorithms to parameterize closed contours while robustly handling the presence of outliers and occlusions in the data. These properties are desirable in computer vision applications where edge point classification algorithms often include outliers similar to the droplet analysis example presented above.

### ACKNOWLEDGMENTS

The author would like to thank Dr. Alison Hoxie from the University of Minnesota Duluth for providing the droplet combustion images that were analyzed. This work was supported by the Swenson College of Science and Engineering at the University of Minnesota Duluth.

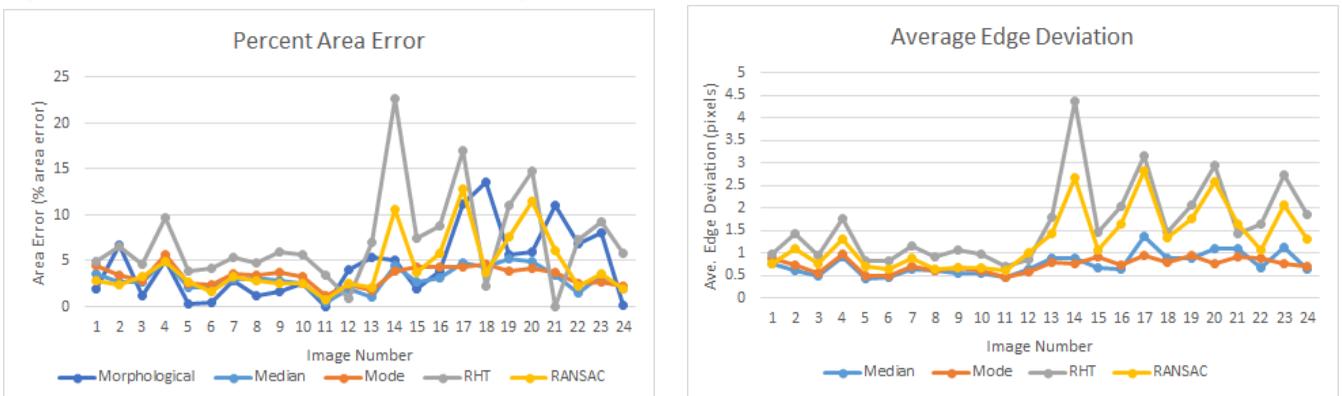

Fig. 16: Fitting error for the robust contour fitting algorithms evaluated.

Table 5: Average area error and average edge deviation error for robust curve fitting algorithms applied to the analysis of droplet combustion images.

|  | Average Percent Area Error | Average Edge Deviation (pixels) |
|---|---|---|
| Median | 3.0 | 0.75 |
| Mode | 3.4 | 0.74 |
| RHT | 7.2 | 1.64 |
| RANSAC | 4.3 | 1.30 |
| Morphological | 4.4 | |